\documentclass[10pt,twocolumn,letterpaper]{article}

\usepackage{iccv}
\usepackage{times}
\usepackage{epsfig}
\usepackage{graphicx}
\usepackage{amsmath}
\usepackage{amssymb}

\usepackage{subcaption}

\usepackage{booktabs}
\usepackage{float}
\usepackage{appendix}
\usepackage{lipsum}
\usepackage{afterpage}
\usepackage{algorithm}
\usepackage{algpseudocode}
\usepackage{mathtools}

\usepackage[normalem]{ulem}

\iccvfinalcopy

\newcommand\blfootnote[1]{%
  \begingroup
  \renewcommand\thefootnote{}\footnote{#1}%
  \addtocounter{footnote}{-1}%
  \endgroup
}

\usepackage[pagebackref=true,breaklinks=true,colorlinks,bookmarks=false]{hyperref}

\usepackage[capitalize]{cleveref}
\crefname{section}{Sec.}{Secs.}
\Crefname{section}{Section}{Sections}
\Crefname{table}{Table}{Tables}
\crefname{table}{Tab.}{Tabs.}

\def\iccvPaperID{2893} 

\ificcvfinal\fi

\begin{document}

\title{ Towards Realistic Generative 3D Face Models }


\author{
Aashish Rai$^{1}$
\and
Hiresh Gupta$^{*1}$
\and
Ayush Pandey$^{*1}$
\and
Francisco Vicente Carrasco$^{1}$
\and
Shingo Jason Takagi$^2$
\and
Amaury Aubel$^2$
\and
Daeil Kim$^2$
\and
Aayush Prakash$^2$
\and
Fernando De la Torre$^1$
\vspace{0.05in}
\and
\centerline{$^1$Carnegie Mellon University \hspace{0.2in} $^2$Meta Reality Labs}
\and
{\tt\small \url{https://aashishrai3799.github.io/Towards-Realistic-Generative-3D-Face-Models}}
}

\vspace{-0.2in}

\twocolumn[{
\renewcommand\twocolumn[1][]{#1}%
\newcommand{\dalle}[0]{DALL$\cdot$E\xspace}
\maketitle
\begin{center}
    \centering
    \captionsetup{type=figure}
    \includegraphics[width=0.95\linewidth]{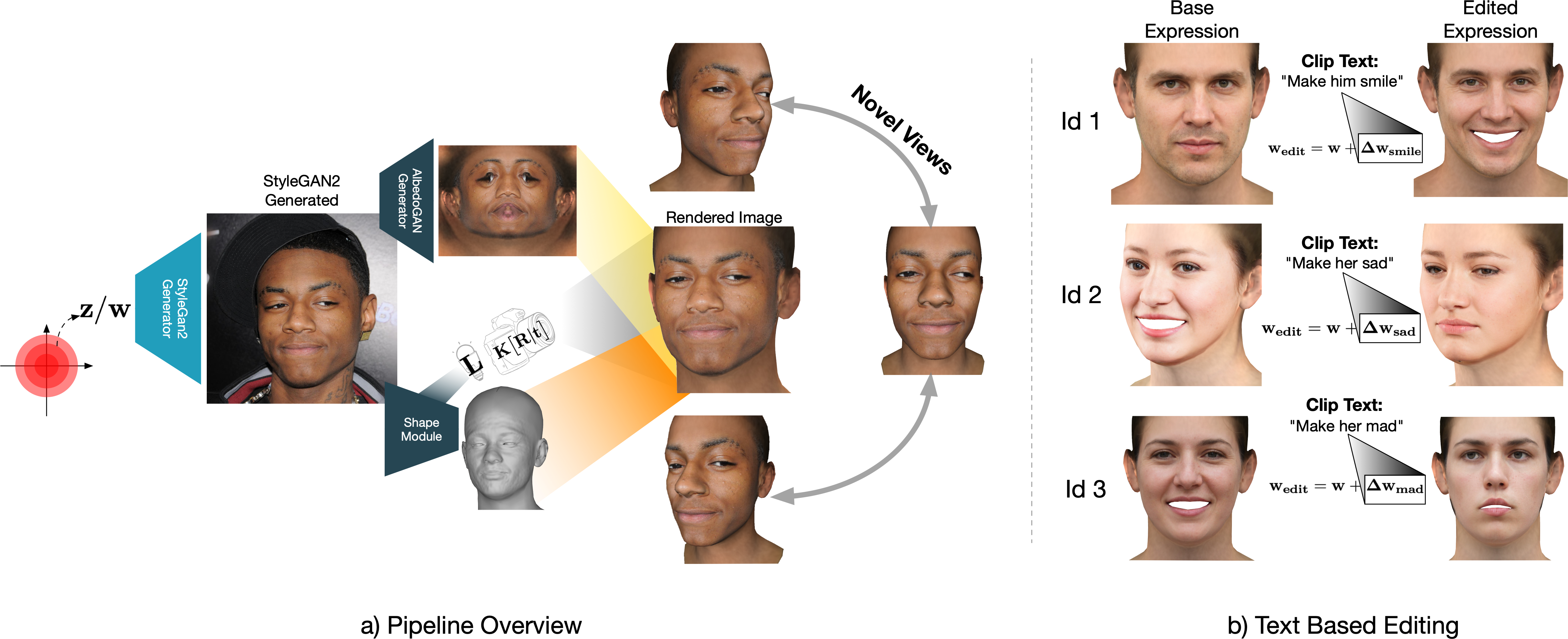}
    \vspace{-0.075in}
    \caption{3D generative face model. a) High-resolution 3D shape and albedo recovered from a StyleGAN2 generated image. Novel views can be rendered using the estimated face model. b) Editing of 3D faces with text. This method allows for 3D expression manipulation through guidance with the CLIP model.}\label{fig:fig1}
\end{center}
}]
\maketitle

\begin{abstract}
\vspace{-0.1in}
\blfootnote{* equal contribution}
In recent years, there has been significant progress in 2D generative face models fueled by applications such as animation, synthetic data generation, and digital avatars. However, due to the absence of 3D information, these 2D models often struggle to accurately disentangle facial attributes like pose, expression, and illumination, limiting their editing capabilities. To address this limitation, this paper proposes a 3D controllable generative face  model to produce high-quality albedo and precise 3D shape leveraging existing 2D generative models. By combining 2D face generative models with semantic face manipulation, this method enables editing of detailed 3D rendered faces. 
The proposed framework utilizes an alternating descent optimization approach over shape and albedo. Differentiable rendering is used to train high-quality shapes and albedo without 3D supervision.
Moreover, this approach outperforms most state-of-the-art (SOTA) methods in the well-known NoW and REALY benchmarks for 3D face shape reconstruction. It also outperforms the SOTA reconstruction models in recovering rendered faces' identities across novel poses by an average of $10\%$. Additionally, the paper demonstrates direct control of expressions in 3D faces by exploiting latent space leading to text-based editing of 3D faces.

\end{abstract}

\vspace{-0.2in}
\section{Introduction}
\vspace{-0.1in}

The success of language models like GPT-3~\cite{radford2019language}, and more recently, the release of text-to-image models like  GLIDE~\cite{Nichol2021GLIDETP}, DALLE-2~\cite{ramesh2022hierarchical}, or Imagen \cite{Saharia2022PhotorealisticTD} have all contributed to the enormous popularity of generative AI. Besides generating images with unprecedented visual quality, these models also show remarkable generalization ability to novel texts with complex compositions of concepts, making them generalists for image synthesis. In the context of faces, StyleGAN2~\cite{karras2020analyzing} has the capability of generating powerful face images that are frequently indistinguishable from reality. While these 2D generative models create high-quality faces for many applications of interest, such as facial animation~\cite{ichim2015dynamic, weise2011realtime}, expression transfer~\cite{kim2018deep, thies2016face2face, olivier2021facetunegan} virtual avatars~\cite{lombardi2018deep}, 
these 2D models often encounter difficulties when it comes to effectively  disentangle facial attributes like pose, expression, and illumination.  As a result, their capacity to edit such attributes is limited. Moreover, 
a  3D representation (shape, texture) is crucial to many entertainment industries—including games, animation, and visual effects— that are demanding 3D content at increasingly enormous scales to create immersive virtual worlds.  Recall that many applications of interest require 3D assets that are consumable by a graphics engine (e.g., Unity \cite{unity}, Unreal \cite{unrealengine}).

To address this demand, recently, researchers have proposed generative models to generate 3D faces~\cite{abrevaya2019decoupled, taherkhani2022controllable, gecer2020synthesizing}. Even though these algorithms perform well, the lack of diverse and high-quality 3D training data has limited the generalization of these algorithms and their use in real-world applications~\cite{toshpulatov2021generative}. Another line of research involves using parametric models like 3DMM\cite{3dmm}, BFM\cite{bfm}, FLAME\cite{3dmm}, and derived methods \cite{deca, occlusion_robust_mofa, deep3dface_recon, tewari, ringnet} to approximate the 3D geometry and texture of a 2D face image. While these 3D face reconstruction techniques can reasonably recover low-frequency details, they typically do not recover high-frequency details. Also, predicting high-resolution texture maps that capture details remains an unaddressed problem. Most of the works focusing in this direction either emphasize mesh or texture. However, a generative 3D face model that can generate both high-quality texture and a detailed mesh with the same quality as 2D models is still missing.  

This paper proposes a 3D generative model for faces using a self-supervised approach that can generate high-resolution texture and capture high-frequency details in the geometry. The method leverages a pretrained StyleGAN2 to generate high-quality 2D faces (see Fig.~\ref{fig:fig1}~a). 
We propose a network, AlbedoGAN, that  generates light-independent albedo directly from StyleGAN2’s latent space. For the shape component, the FLAME model~\cite{flame} is combined with per-vertex displacement maps guided by StyleGAN's latent space, resulting in a higher-resolution mesh. The two networks for albedo and shape are trained in alternating descent fashion. The proposed method outperforms SOTA methods in shape estimation, such as DECA~\cite{deca} and MICA~\cite{MICA:ECCV2022}, by $20\%$ and $1.1\%$, respectively. It's worth noting that MICA only generates a neutral and frontal smooth mesh, while the proposed algorithm can generate any expression. Fig.~\ref{fig:fig1}(a) shows how an image generated by StyleGAN2 can be uplifted to 3D with a detailed shape and albedo, being able to render realistic 3D faces.  Finally, given the 3D face asset, our algorithm can edit the face in 3D. For example, Fig.~\ref{fig:fig1}(b) illustrates expression manipulation through text-based editing guided by the CLIP model~\cite{radford2021learning}. Briefly stated, our main contributions are:

1. A self-supervised method to leverage StyleGAN2 into a 3D generative model producing high-quality albedo and a detailed mesh.  We introduce AlbedoGAN, a single-pass albedo prediction network that generates high-resolution albedo and decouples illumination using shading maps. 

2. We show that our model outperforms existing methods in capturing high-frequency facial details in a mesh.  Moreover, the proposed method reconstructs 3D faces that recover identity better than SOTA methods. 

3. We propose a displacement map generator capable of decoding per-vertex displacements directly from StyleGAN's latent space using detailed normals of the mesh.

4. Since our entire architecture can generate 3D faces from StyleGAN2’s latent space, we can perform face editing directly in the 3D domain using the latent codes or text. 


\vspace{-2mm}
\section{Related Work}
\vspace{-1mm}

\label{sec:formatting}

Reconstructing a 3D Face from a single 2D image is an ill-posed problem that has intrigued researchers for decades. Blanz et al. \cite{3dmm} took the first significant step in this direction when they introduced 3D Morphable Models (3DMM) \cite{3dmm} in 1999 as general face representation. Their work inspired decades of work in estimating parameters for 3DMM to find a textured mesh that best fits an input 2D face. While estimating parameters for parametric models like 3DMM and its advanced versions like FLAME \cite{flame} has been the bulk of the focus for researchers, there has also been a good amount of work done in learning volumetric representations (e.g., NeRF) for a face. In particular, the focus has been on using Neural Implicit Representation \cite{neural-head-avatars, nerface, imavatar, lolnerf, pi_gan, mofa_nerf} to learn density and radiance to represent a face. However, it has been widely noticed that they are prone to generating artifacts and consume a lot of time to render these detailed representations. Furthermore, these methods do not generate a topologically uniform mesh, and therefore do not directly serve applications in graphic engines, face animation, avatar creation, etc. Due to the above-mentioned reasons, we do not consider implicit representation in our research. In the following sections, we describe work that is related to individual components of our framework.

\begin{figure*}[!ht]
\centering
\includegraphics[width=5.in]{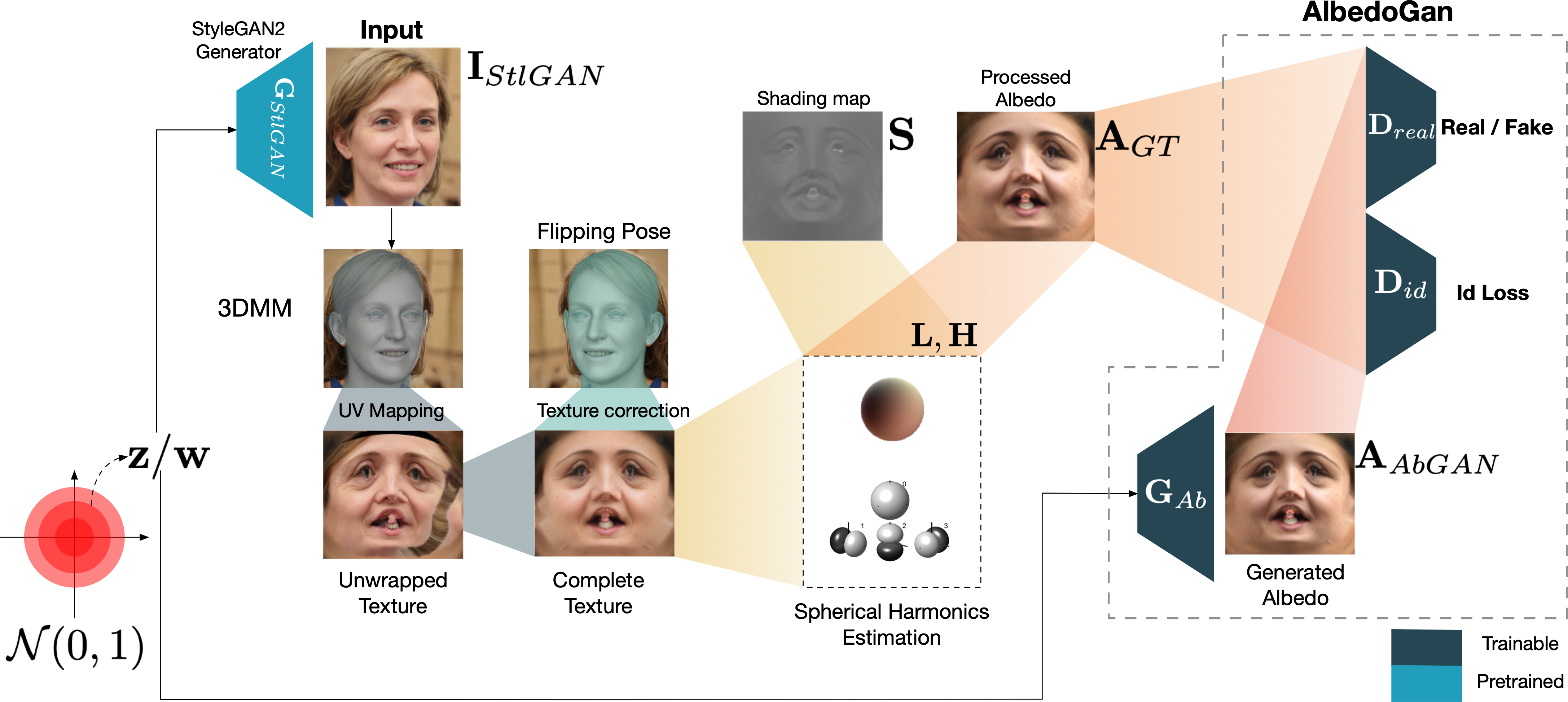} 
\caption{AlbedoGAN. Pose-invariant albedo, $\textbf{A}_{GT}$, obtained by texture extraction and synthesis \ref{sec:tex_extraction}, is used to train StyleGAN2 generator, $\textbf{G}_{Ab}$, for a given latent code $\textbf{z}/\textbf{w}$. We use a 3DMM fitting, image blending, and Spherical Harmonics to extract $\textbf{A}_{GT}$. $\textbf{D}_{real}$ and $\textbf{D}_{id}$ are introduced to generate realistic images and identity consistent albedo, respectively.}
\label{fig:tex_prep}
\vspace{-2mm}
\end{figure*}


\vspace{-1mm}
\subsection{Texture Generation for 3D Mesh}

Most of the work done in synthesizing texture for a mesh can be broadly divided into two parts: using a parametric texture model like 3DMM or Basel Face Model (BFM) \cite{deca, deep3dface_recon, occlusion_robust_mofa, tewari} or a GAN-based approach \cite{Gecer_2021_CVPR, gecer2020tbgan, Gecer_2019_CVPR, fastganfit, textureGeneration, texture-completion-iccv, gecer2019ganfit, uvgan} to generate texture. 
Using a parametric model BFM \cite{bfm}, works like \cite{deca, deep3dface_recon, occlusion_robust_mofa, tewari} learn an encoder to predict the parameters that generate a texture that best fits the visible part of the face. 
Since the model uses representation in low dimensional PCA-based space, they generate an approximate texture that often lacks high-frequency details, which correspond to low variation direction in the projected latent space and do not lead to photo-realistic rendering. Recently, with the advent of GANs, there are works that leverage them to extract texture \cite{Gecer_2021_CVPR, gecer2020tbgan, texture-completion-iccv}. TBGAN \cite{gecer2020tbgan} trains a Progressive GAN to generate a high-quality texture with a differentiable renderer in a supervised setting. On the other hand, OSTeC \cite{Gecer_2021_CVPR} uses 3DMM as initialization to generate a textured mesh and then uses GAN inversion with StyleGAN2 to generate multiple views of this mesh and extract texture. This is an extremely time-consuming and takes several minutes per image. We propose AlbedoGAN, which is able to generate high-quality texture in a single pass in a time-efficient manner. AlbedoGAN generates textures that maintain identity over multiple poses - where most previous methods struggle.





\subsection{3D Shape prediction from 2D Image}


 Similar to texture, Blanz et al.'s seminal work \cite{3dmm} can represent a face mesh in a low-dimensional PCA-based space. This led to the development of a huge corpus of work \cite{deep3dface_recon, tewari, ringnet, occlusion_robust_mofa} in 3D face reconstruction focused on learning an encoder that could predict parameters for generating shapes using 3DMM, given a 2D image. The encoder could be learned in a self-supervised way with 2D image losses \cite{deca, occlusion_robust_mofa, deep3dface_recon, tewari, ringnet}, or with 3D supervision, \cite{MICA:ECCV2022}. After 3DMM\cite{3dmm}, there have been newer parametric models BFM \cite{bfm} and FLAME \cite{flame}, which have been learned from more subjects and encode the structure of the face better. The explosive development of 3D Face Reconstruction led to the creation of NoW benchmark \cite{ringnet} as a common ground for comparison across 3D Face reconstruction methods. Currently, DECA \cite{deca}, and MICA \cite{MICA:ECCV2022} show best reconstruction on NoW Benchmark. DECA's architecture is inspired by RingNet \cite{ringnet}. However, instead of 3DMM, DECA uses FLAME \cite{flame}, and it adds an encoder-decoder network that learns to generate displacement maps to learn animatable details in UV space. MICA \cite{MICA:ECCV2022}, the current state-of-art model, leverages ArcFace backbone \cite{deng2018arcface} to learn the actual shape of the face by regressing it on a high-quality 3D face scan dataset\cite{LYHM, stirling, facewarehouse}.

\section{AlbedoGAN Training}


Albedo constitutes one of the crucial parts of a 3D face model, since face appearance is largely dictated by it. To generate high quality 3d models, we need to generate albedo that generalize over pose, age, and ethnicity. However, training such a diverse albedo generative model requires a massive database of 3D scans, which is neither cost nor time effective. An efficient way of extracting textures from existing 2D images is fitting a 3DMM and capturing a UV mapped texture. However, this "pseudo" texture does not generalize well over poses nor disentangle shadows. In this paper, we leverage 3DMM fitting, image blending, and Spherical Harmonics lighting to capture high-quality $1024\times1024$ resolution albedo that generalizes well over different poses and tackles shading variations.

This section describes albedo training, refer Fig.~\ref{fig:tex_prep} for an overview.  The first step includes texture extraction and correction, \ref{sec:tex_extraction}, followed  by the use of a spherical harmonics model to extract albedo from texture (Section~\ref{sec:sh_model}). Section~\ref{sec:pretraining_albedo} explains training AlbedoGAN - a StyleGAN2 model to generate albedo corresponding to the given latent code $\textbf{w} \in \mathbb{R}^{18 \times 512}$. 





\subsection{Texture Extraction and Correction}
\label{sec:tex_extraction}


First, we establish a correspondence between the input 2D image $\textbf{I} \in \mathbb{R}^{w \times h \times 3}$ and the UV domain by taking orthogonal projection of a mesh fitted on the given image using a 3DMM \cite{flame}. Using this correspondence, we get the RGB values for the UV texture and perform barycentric interpolation to fill out the missing pixels. 

Now, texture correction is performed to fill the occluded areas by leveraging pose information. This happens by projecting the flipped input image and fitted mesh, and collecting the pixels corresponding to the missing parts. These pixels are blended to the original texture to get a complete pose-invariant texture.





\subsection{Albedo from Texture}
\label{sec:sh_model}

As shown in Fig. \ref{fig:tex_prep}, the next step includes obtaining an albedo and shading map from the unevenly illuminated texture. Following previous works \cite{deca, deep3dface_recon},  we made the following assumptions: (1) The illumination model is Spherical Harmonics (SH), (2) light source is distant and monochromatic, and (3) surface reflectance is Lambertian. The shading map can thus be calculated as 

\vspace{-2mm}

    \[ \textbf{S}_{ij} = \sum_{b=1}^{9} \textbf{L}_b \textbf{H}_b (\textbf{N}_{ij}) \]

where, $\textbf{H}_b : \mathbb{R}^3 \rightarrow \mathbb{R}$ are the SH basis function, $\textbf{L}_b \in \mathbb{R}^3$ are SH coefficients, and $N_{ij} \in \mathbb{R}^3$ are surface normals. The relation between albedo $\textbf{A} \in \mathbb{R}^{w \times h \times 3}$, texture $\textbf{T} \in \mathbb{R}^{w \times h \times 3}$, and shading map $\textbf{S} \in \mathbb{R}^2$ can then be defined as  
$\textbf{T}_{ij} = \textbf{A}_{ij} \odot \textbf{S}_{ij}$, where, $\odot$ is the Schur product.


\begin{figure*}[!ht]
\centering
\includegraphics[width=7. in]{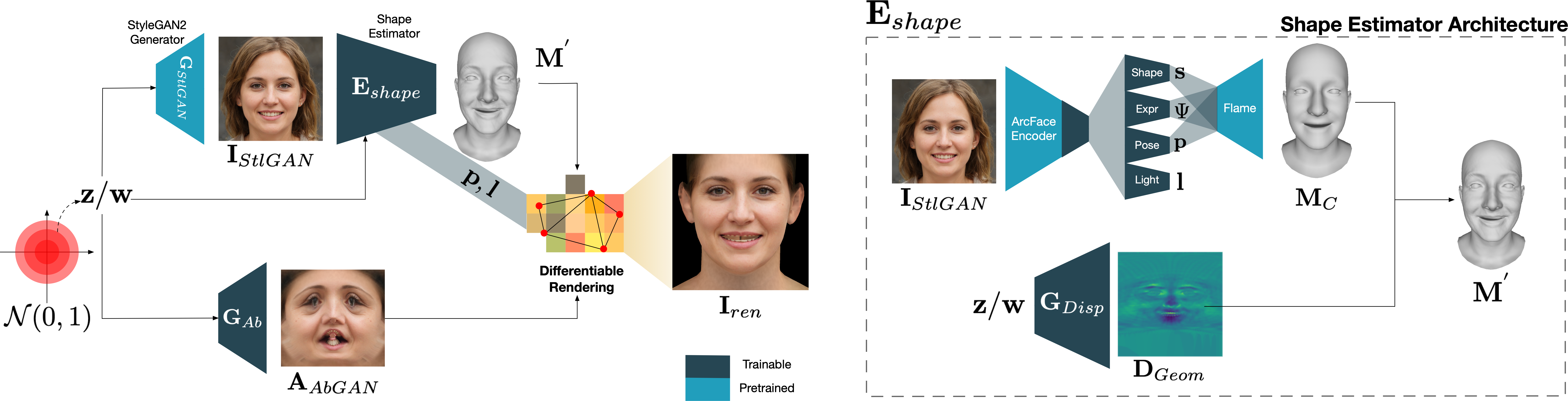}
\vspace{1mm}
\caption{Overview of our generative model. The AlbedoGAN generator, $\textbf{G}_{Ab}$, is used to synthesize albedo $\textbf{A}_{AbGAN}$ corresponding to a latent code $\textbf{w}$. $\textbf{G}_{StlGAN}$ generates a 2D image, $\textbf{I}_{StlGAN}$, given to the shape model, $\textbf{E}_{shape}$, to get a detailed mesh, \textbf{M}'.
Finally, a differentiable renderer (DR) is used to synthesize $\textbf{I}_{ren}$ from the mesh $\textbf{M}'$, albedo $\textbf{A}_{AbGAN}$, lighting $\textbf{l}$, and pose $\textbf{p}$.
Losses between $\textbf{I}_{ren}$ and $\textbf{I}_{StlGAN}$ are used to train the shape model and the AlbedoGAN via Alternating Descent.}
\vspace{-1mm}
\label{fig:architecture}
\vspace{-1mm}
\end{figure*}

\subsection{Training}  
\label{sec:pretraining_albedo}

In this section, we address the AlbedoGAN model training procedure. Later, the resulting model will be fine-tuned during the shape and displacement map training process, taking into account geometry and more sophisticated Phong illumination model.




Our approach, AlbedoGAN, is built upon a generative model that can synthesize face images corresponding to a latent vector, to this end we selected StyleGAN2. 
This model is best suited to our requirements as it uses a mapping network $M_{L}$ that maps an input noise vector $\textbf{z} \in \mathbb{R}^{512}$ to an intermediate latent vector $\textbf{w} \in \mathbb{R}^{18 \times 512}$. This mapping, $\textbf{w} = M_{L}(\textbf{z})$, adds the ability for manipulation and better projection. Consequently, we use $\textbf{w}$ as latent space for AlbedoGAN.

Hence, face images are generated by randomly sampling $\textbf{w}$ using a pretrained StyleGAN2. The generated images act as input, $\textbf{I}_{StlGAN}$, see Fig.\ref{fig:tex_prep}. The same latent codes, $\textbf{w}$,  are used in the AlbedoGAN generator, to produce $\textbf{A}_{AbGAN}$. As shown in Fig.\ref{fig:tex_prep}, $\textbf{I}_{StlGAN}$ is passed through the texture extraction, \ref{sec:tex_extraction} and albedo extraction,  \ref{sec:sh_model}, steps to obtain $\textbf{A}_{GT}$. This albedo is used as a ground truth for the training of the real/fake discriminator, $\textbf{D}_{real}$. Additionally, we constraint AlbedoGAN to generate identity consistent albedos by introducing an identity discriminator, $\textbf{D}_{id}$. To this intent, we use the features of a pretrained face recognition model \cite{he2016deep} $F : \mathbb{R}^{w' \times h' \times 3} \rightarrow \mathbb{R}^{512}$. Our identity loss is defined as cosine distance between the identity features of predicted albedo $\textbf{A}_{AbGAN}$ and $\textbf{A}_{GT}$ as: 


\vspace{-2mm}
\begin{equation}
 L_{id} (\textbf{A}_{AbGAN}, \textbf{A}_{GT}) = 1 - \frac{F(\textbf{A}_{AbGAN}) . F(\textbf{A}_{GT})}{||F(\textbf{A}_{AbGAN})||_2 ||F(\textbf{A}_{GT})||_2} 
\label{eq:l_id}
\end{equation}


\section{Alternating Descent in Albedo and Shape}

In this section, we describe our regression method to 3D shape given a face image and the Differentiable Rendering, DR, based approach to fine-tune AlbedoGAN. This fine-tuning process for AlbedoGAN takes into account expression, camera pose, and the Phong illumination model. 

Unfortunately, jointly optimizing all the components (shape, albedo, illumination, etc.) that produce the best rendered face that is consistent with the input image is computationally expensive. Thus, we propose using Alternating Descent for optimization. First, we optimize the shape for a few iterations while freezing AlbedoGAN. Next, AlbedoGAN is fine-tuned using the updated 3D shape, with more detailed normals of the shape. This alternating optimization cycle is repeated throughout the course of the training process. Next, we first describe albedo optimization, \ref{sec:albedo_optimization}, followed by shape optimization,  \ref{sec:shape_optimization}.

\subsection{Albedo optimization}
\label{sec:albedo_optimization}
To fine-tune AlbedoGAN using the information of the 3D shape and illumination model, we first assume we have an estimate of a detailed 3D shape $\textbf{M}'$. As shown in Fig.~\ref{fig:architecture}, given an estimated mesh $\textbf{M}'$, predicted albedo $\textbf{A}_{AbGAN}$, pose $\textbf{p}$, and light $\textbf{l}$, we can generate a detail rendered image $\textbf{I}_{ren}$ using DR, $R$ as:
\[ \textbf{I}_{ren} = R(\textbf{M}', \textbf{A}_{AbGAN}, \textbf{p}, \textbf{l}) \]  


The overall loss function $\mathcal{L}$ is defined as a sum of the following terms: 

\vspace{-4mm}

\[ 
\mathcal{L} = \lambda_{sym\_rec} \mathcal{L}_{sym\_rec} + 
\lambda_{id} \mathcal{L}_{id} + 
\lambda_{perc} \mathcal{L}_{perc} + 
\lambda_{lmk} \mathcal{L}_{lmk} 
\]

Where each loss is defined as follows:



{\bf Symmetric Reconstruction Loss, $\mathcal{L}_{sym\_rec}$:} A simple supervision function that encourages low-level similarity in the predicted image and the corresponding ground truth and symmetry in the estimated albedo. We use the Mean Squared Error (MSE) to calculate reconstruction error.
\begin{equation*}
\begin{multlined}
 \mathcal{L}_{sym\_rec} (\textbf{I}_{ren}, \textbf{I}_{StlGAN}) = \\
 \textrm{MSE}(\textbf{I}_{StlGAN}, \textbf{I}_{ren}) + \underbrace{\textrm{MSE}(\textbf{I}^{'}_{StlGAN}, \textbf{I}^{'}_{ren})}_{\text{ Albedo symmetry consistency term}}   
\end{multlined}
\end{equation*}
\label{eq:l_rec}

\vspace{-2mm}

Where $\textbf{I}^{'}_{StlGAN}$ is the flipped ground truth obtained through StyleGAN2, and $\textbf{I}^{'}_{ren}$ is the rendered estimated image through flipping $\textbf{A}_{AbGAN}$, pose $\textbf{p}$, and light $\textbf{l}$. 

{\bf Identity Loss, $\mathcal{L}_{id} (\textbf{I}_{ren}, \textbf{I}_{StlGAN})$:} This loss term is introduced with the intent of making the AlbedoGAN generator learn to match the identity of the rendered face, $\textbf{I}_{ren}$, with the ground truth, $\textbf{I}_{StlGAN}$. As in section \ref{sec:pretraining_albedo}, we use a pretrained face recognition model \cite{he2016deep} for feature extraction. The cosine distance, eq.~\ref{eq:l_id}, is used to calculate the identity loss while fine-tuning the model.

{\bf Perceptual Loss, $\mathcal{L}_{perc}$:} This perceptual loss is introduced with the goal of  forcing AlbedoGAN to generate a $\textbf{A}_{AbGAN}$ that matches the visual appearance of $\textbf{I}_{StlGAN}$. Motivated by the existing research, we selected a VGG16 based feature extractor \cite{schroff2015facenet}, a pretrained face recognition model. We use the output of $relu3\_3$ as the image features. The loss is calculated by the L2 distance between the feature vectors from $\textbf{I}_{ren}$ and $\textbf{I}_{StlGAN}$.  


{\bf Landmark Loss, $\mathcal{L}_{lmk}$:} AlbedoGAN is also fine tuned-using using 68 facial landmarks detected on the ground truth and the rendered image to avoid misaligned generations. We used a SOTA face landmark detection~\cite{yin2020fan} to predict 68 landmarks on $\textbf{I}_{StlGAN}$ and $\textbf{I}_{ren}$. The loss is calculated using MSE between the two set of landmarks. 


\subsection{Shape Model and Optimization}
\label{sec:shape_optimization}
We proceed with the shape optimization, while freezing AlbedoGAN. We sample a latent vector $\textbf{w} \in \mathbb{R}^{18 \times 512}$ and use a pretrained StyleGAN2 model, $\textbf{G}_{StlGAN}$, to generate a 2D face image, $\textbf{I}_{StlGAN}$, and AlbedoGAN generator, $\textbf{G}_{AbGAN}$, to generate albedo, $\textbf{A}_{AbGAN}$. 
Figure~\ref{fig:architecture} describes the detailed architecture of our shape model, $\textbf{E}_{shape}$. We leverage ArcFace backbone \cite{deng2019arcface} to predict the face shape ($\textbf{s}$), expression ($\psi$), lighting ($\textbf{l}$), and camera pose ($\textbf{p}$) parameters for the given image $\textbf{I}_{StlGAN}$. 
The shape  embedding vector $\textbf{s}\in \mathbb{R}^{300}$, pose $\textbf{p} \in \mathbb{R}^6$, and expression $\psi \in\mathbb{R}^{50}$ parameters are fed into a parametric face model that gives us a coarse mesh representation ($\textbf{M}_c$) as described below: 
\begin{equation}
 \textbf{M}_c(s, \psi) = \textbf{T} + \textbf{B}_S \textbf{s} + \textbf{B}_\psi \psi
\label{eq:coarse_mesh}
\end{equation}

where $\textbf{M}_c$ represents the generated coarse mesh synthesized by a 3DMM decoder. Specifically, we use FLAME \cite{flame} as our mesh decoder, which generates a coarse mesh with $N=5023$ vertices. This coarse mesh is computed by using a template mesh, $\textbf{T}\in \mathbb{R}^{3N}$, representing a mean human face and different principal components $\textbf{B}_s \in \mathbb{R}^{3N\times300}$ and $\textbf{B}_\psi \in \mathbb{R}^{3N\times50}$ corresponding to the shape and expression terms respectively.

To capture high-frequency details in meshes, we learn a displacement generator $\textbf{G}_{Disp}$ to augment the coarse mesh, $\textbf{M}_c$, with a detailed UV displacement map $\textbf{D}_{Geom} \in [-0.01, 0.01]^{n \times n}$. Recent research \cite{latent_space2, latent_space1} has shown that StyleGAN's latent space contains information about high-frequency details of a face. Using this insight, we predict displacement maps to capture expression and pose dependent per vertex offsets. The latent code $\textbf{w}$ is the same as used in AlbedoGAN and the StyleGAN2 model. Finally, we combine the displacement map along the vertex normals of the mesh $\textbf{M}_c$ to get a detailed mesh $\textbf{M}'$ by adding them in the UV domain. 


We use the detailed mesh, $\textbf{M}'$, along with the predicted pose $\textbf{p}$, light $\textbf{l}$ parameters, and the synthesized albedo $\textbf{A}_{AbGAN}$ to render an image $\textbf{I}_{ren}$ as described below:

\begin{equation}
\textbf{I}_{ren} = R(\textbf{M}_c, \textbf{A}_{AbGAN}, \textbf{p}, \textbf{l})
\label{eq:rendered_mesh}
\end{equation}

We apply multiple 2D image-based losses, including identity loss, perceptual loss, and landmark loss between $\textbf{I}_{StlGAN}$ and $\textbf{I}_{ren}$ to optimize the mesh representation in a self-supervised fashion. 
In addition to the losses, we also calculate a shape center loss eq.~\ref{eq:shape_consistency} on images belonging to identity $i$. In particular, eq.~\ref{eq:shape_consistency} tries to reduce the distance between shape vector for all the images and their corresponding mean $\mu_i$.
Besides this, we also perform L2 regularization, eq.~\ref{eq:l_reg}, on predicted shape $\textbf{s}$, expression $\psi$, and displacement maps $\textbf{D}_{Geom}$ that enforce a prior distribution towards the mean face.

\begin{equation}
\mathcal{L}_{sc}=\sum_{i=0}^{N}\sum_{k=0}^{K}\left\| \textbf{s}_{i,k}-\mu_i \right\|^2_2
\label{eq:shape_consistency}
\end{equation}

\begin{equation}
\mathcal{L}_{reg} = {\lVert \textbf{s} \rVert}^2_2 + {\lVert \psi \rVert}^2_2 + {\lVert \textbf{D}_{Geom} \rVert}^2_F
\label{eq:l_reg}
\end{equation}
The overall loss function $\mathcal{L}$ is defined as a weighted sum:
\[ 
\mathcal{L} = \lambda_{id} \mathcal{L}_{id} + 
\lambda_{perc} \mathcal{L}_{perc} + 
\lambda_{lmk} \mathcal{L}_{lmk} + 
\lambda_{sc} \mathcal{L}_{sc} +
\lambda_{reg} \mathcal{L}_{reg}
\]

\section{Experiments}

This section describes the implementation details, quantitative, and qualitative evaluation of the shape and texture reconstruction models, along with SOTA comparison. 




\subsection{Dataset}

We randomly sampled $100K$, $512$-dimensional random vectors $\textbf{z} \in \mathbb{R}^{512}$ from a Gaussian distribution and generated the corresponding $\textbf{w} \in \mathbb{R}^{18 \times 512}$ from the StyleGANs mapping network as $\textbf{w} = M_{L}(\textbf{z})$. These intermediate latent vectors $\textbf{w}$ are used to generate $100K$ images $\in \mathbb{R}^{1024 \times 1024 \times 3}$ from a pretrained StyleGAN2 \cite{stylegan2} generator. To ensure diversity in the generated images across ethnicity, expression, age, and pose; we followed the work in \cite{rai2021improved}. The texture-preprocessing step (as described in Sec. \ref{sec:tex_extraction}) is used to get the complete GT-albedo corresponding to all the samples in the dataset. 

To train our shape model with 2D images, we chose $30K$ of the previously sampled $\textbf{z}$ vectors. Then we perform latent space editing in the $\textbf{w}$ space to generate a total of $11$ images (belonging to different expressions and poses) per identity using StyleGAN2 implementation \cite{rai2021improved} of InterFaceGAN \cite{shen2020interfacegan}. We estimate $68$ landmarks on all the GT images using the FAN~\cite{yin2020fan} landmark detection.   

\subsection{Implementation Details}

{\bf Albedo Generation:} PyTorch \cite{paszke2019pytorch} is used as the implementation framework on CUDA enabled system with NVIDIA RTX $A4500$ GPUs. We use the PyTorch implementation of StyleGAN2 \footnote{https://github.com/rosinality/stylegan2-pytorch} for albedo and image generation. To generate face images from StyleGAN2, we use the official pretrained weights trained on the FFHQ dataset for $1024 \times 1024$ resolution. We use Adam optimizer to train the AlbedoGAN with learning rate $\alpha_{gen} = \alpha_{disc} = 2e^{-3}$ and $\beta_{1} = 0, \beta_{2}=0.99$. The generator was regularized after every $4$ iteration, while the discriminator after every $16$ training iterations. 

In the first step, we train the AlbedoGAN from GT-albedo and use the GAN loss and ID loss eq.\ref{eq:l_id} for supervision. Similar to StyleGAN2, for the GAN loss, we use the element-wise Softplus, which can be defined as $Softplus(x) = \frac{1}{\beta} * \log(1 + \exp^{(\beta * x)})$.  The $\mathcal{L}_{ID}$ is calculated between the predicted albedo from the generator $\textbf{A}_{AbGAN}$ and the GT-albedo $\textbf{A}_{GT}$. The $\lambda_{ID}$ was set to $1$ during this training. We trained the model on $8$ GPUs, and it took around $32$ hours for complete training (batch size $32$).

This gave us a robust albedo generator capable of generating a pose-invariant albedo corresponding to a given $\textbf{z} \in \mathbb{R}^{512}$ or $\textbf{w} \in \mathbb{R}^{18 \times 512}$. 

\begin{figure}[t]
\centering
\includegraphics[width=0.42\textwidth]
{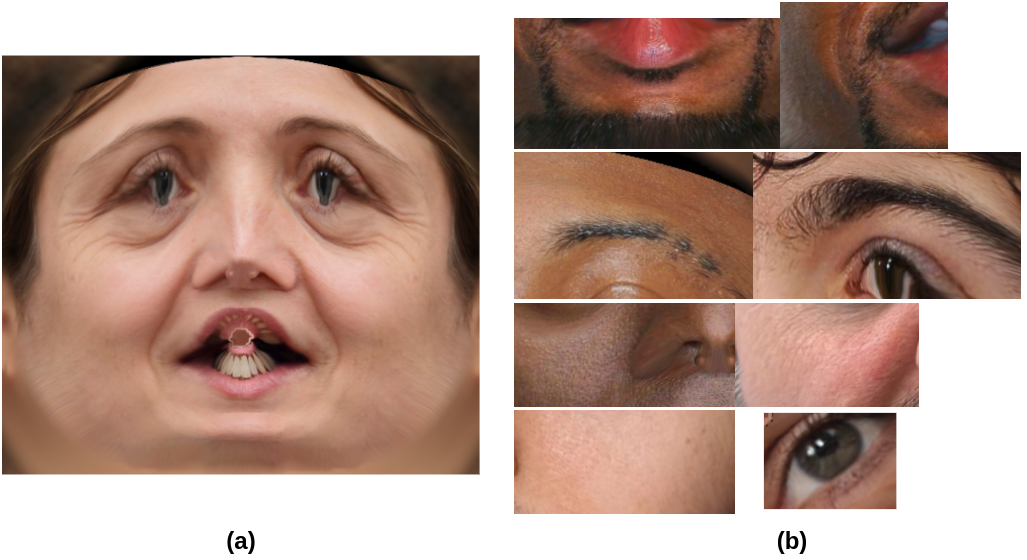}

\caption{(a) Randomly generated albedo from AlbedoGAN. (b) Patches of randomly generated albedo. AlbedoGAN can generate high-quality albedo of $1K$ resolution.}
\label{fig:tex_samples}
\vspace{-3mm}
\end{figure}

{\bf Optimizing Shape:} We train our shape model, $\textbf{E}_{shape}$, on the synthetically generated images by StyleGAN2 capturing multiple images of the same identity across varying expression \& pose. We run a face detector \cite{guo2021sample} on the input images and scale the face crops to a resolution of $224 \times 224$ before passing them to our shape model. The shape model consist of an ArcFace backbone that is initialized to the weights learned by \cite{MICA:ECCV2022} and a convolution-styled decoder ($\textbf{G}_{Disp}$) respectively. The whole pipeline is optimized using Adam Optimizer with a learning rate of $1e^{-4}$. The final loss is calculated between rendered images $\textbf{I}_{ren}$ and GT $\textbf{I}_{StlGAN}$, where $\lambda_{id}$ is set to $0.5$, and  $\lambda_{perc}$, $\lambda_{lmk}$, $\lambda_{sc}$ and $\lambda_{reg}$  are set to $1, 5, 1$ and $1e^{-4}$ respectively. 

\begin{figure}[t]
\centering
\includegraphics[width=0.40\textwidth]{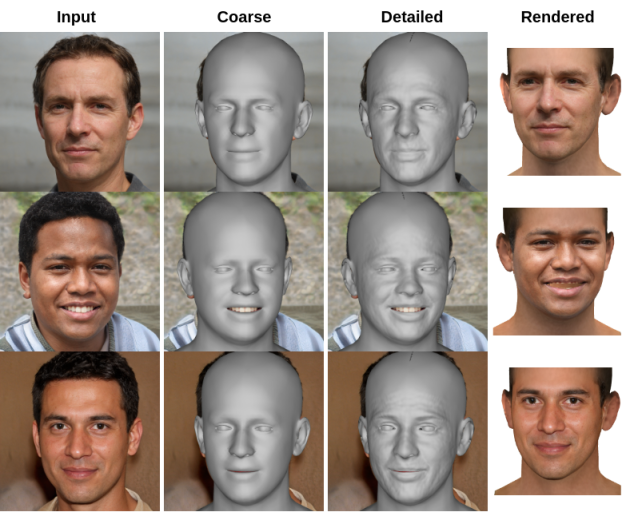}
\vspace{-1mm}
\caption{Randomly generated coarse mesh, detailed mesh and rendered faces from our model, for input 2D faces.}
\label{fig:stylegan3d_samples}
\vspace{-4mm}
\end{figure}

{\bf Fine-tuning AlbedoGAN using DR:} Once we have a pretrained AlbedoGAN and a good shape estimator, we now fine-tune the AlbedoGAN. This makes the albedo generator learn to capture more details from the GT-face. PyTorch3D is used as the differentiable renderer in all our experiments under this section. We kept using $\mathcal{L}_{gan}$ from previous AlbedoGAN training but gave more gravity to the rendering losses. The rendering loss is calculated between $\textbf{I}_{StlGAN}$ and $\textbf{I}_{ren}$ along with a symmetric reconstruction loss between $\textbf{A}_{GT}$ and $\textbf{A}_{AbGAN}$ to maintain low-level consistency in the fine-tuned albedo. $\lambda_{sym\_rec}$ and $\lambda_{ID}$ were set to $0.1$ for albedo based losses. $\lambda_{sym\_rec}$, $\lambda_{ID}$, $\lambda_{perc}$ and $\lambda_{lmk}$ were set to $1, 1, 1, 0.5$ for rendering losses, respectively. We fine-tuned the model for another $48$ hours, keeping the batch size of $8$. The results of randomly generated textures from our AlbedoGAN are shown in Fig. \ref{fig:tex_samples}. 

Once we fine-tune the AlbedoGAN, we again train the shape model with updated albedo weights and repeat this step multiple times until we get a final model that can synthesize realistic-looking 3D faces corresponding to the given 2D images.

Fig. \ref{fig:fig1}(a), \ref{fig:stylegan3d_samples}, \ref{fig:mesh_comparison} shows the reconstructed 3D faces generated from our model for multiple poses. It is interesting to see how our model generalizes well over different poses and generates realistic-looking 3D faces. Section B.3 in supplementary demonstrates lighting and pose control in rendered faces using corresponding parameters and shading maps. Some additional results on testing our pipeline on real-world images using GAN inversion can be seen in supplementary section C.






\subsection{Evaluation of Shape and Texture}

\subsubsection{NoW Benchmark - shape reconstruction}

\vspace{-1mm}

NoW benchmark \cite{ringnet} is a standard benchmark to evaluate the accuracy of 3D meshes estimated from 2D images. It consists of $2054$ images for $100$ test subjects across different expressions, poses, and occlusions, split across two sets for validation (20 subjects) and test (80 subjects). NoW provides 3D ground truth meshes for each test subject, and the predicted mesh is rigidly aligned with the ground truth mesh using 3D face landmarks. The per-vertex error is then used for all the subjects to compute the mean, median, and standard deviation of the errors. Table \ref{table:recons_error} depicts the comparison of our model with the current SOTA methods, including DECA and MICA.
Fig.~\ref{fig:mesh_comparison} illustrates the visual comparison among these methods.
Our model outperforms the DECA model by achieving a \textbf{23\%} better median error in coarse mesh and a \textbf{20\%} better median error in the detailed mesh.
Our approach can reconstruct realistic-looking rendered faces, and model accurate head shapes, especially for faces with big heads. 
We also observe an improvement over the MICA model that was trained on 3D face scans \cite{LYHM, stirling, facewarehouse} with 2300 subjects on the NoW validation set. As illustrated in \ref{fig:mesh_comparison}, our method produces a more detailed mesh, capturing wrinkles, expression, pose and head shape correctly by only training on synthetic images.

\begin{table}[h]
\centering
\caption{Reconstruction error on the NoW Benchmark.}
\vspace{-2mm}
\fontsize{8}{7}
\small
\scalebox{0.85}{\begin{tabular}{c|c|c|c}
Method            & Median (mm) & Mean (mm) & Std (mm)  \\
\hline
\textbf{Validation Set} & & & \\

Deep3D \cite{deep3dface_recon} & 1.286       & 1.864     & 2.361     \\
DECA \cite{deca} & 1.178       & 1.464     & 1.253     \\
MICA \cite{MICA:ECCV2022}   & 0.913       & 1.130      & \textbf{0.948}     \\
Ours   & \textbf{0.903}       & \textbf{1.122}     & 0.957     \\
\hline

\textbf{Test Set} & & & \\

Deep3D \cite{deep3dface_recon} & 1.11       & 1.41     & 1.21     \\
DECA \cite{deca} & 1.09       & 1.38     & 1.18     \\
Ours   & 0.97       & 1.21     & 1.02     \\
MICA \cite{MICA:ECCV2022} & \textbf{0.90}  & \textbf{1.11}  & \textbf{0.92}     \\
\hline

\label{table:recons_error} 
\end{tabular}}
\end{table}

\subsection{REALY 3D Benchmark}

We also evaluated our method on the most recent REALLY benchmark~\cite{REALY} for single-image 3D face reconstruction from frontal and side view images.  Our results, as presented in Tables~\ref{table:realy_front}, demonstrate a significant improvement over DECA by $\textbf{15\%}$ and MICA by $\textbf{18\%}$. Our method also stands in the \textbf{top 3} (out of 18 methods) on the REALLY benchmark challenge outperforming most of the existing methods. The only two methods better than ours are HRN \cite{HRN} and Deep3D \cite{deep3d_rebuttal}.  However, they both generate only frontal mesh, while our method generates a complete head model and is trained in an {\bf unsupervised} setting.

\begin{table}[h]
      \caption{Single image reconstruction error on REALY Benchmark for Frontal and Side-view images (lower is better).}
      \label{table:realy_front}
    \fontsize{8}{7}
    \small
    \scalebox{0.85}{\begin{tabular}{|l|l|l|l|l|l|}
    \hline
       Method  & @nose & @mouth & @forehead & @cheek & @all \\ \hline\hline

       \textbf{Front View} & & & & & \\

        HRN & 1.722 & 1.357 & 1.995 & 1.072 & 1.537 \\ 
        Deep3D & 1.719 & 1.368 & 2.015 & 1.528 & 1.657 \\  
        \textbf{Ours} & 1.656 & 2.087 & 2.102 & 1.141 & \textbf{1.746} \\  
        GANFit & 1.928 & 1.812 & 2.402 & 1.329 & 1.868 \\ 
        DECA & 1.697 & 2.516 & 2.394 & 1.479 & 2.010 \\ 
        PRNet & 1.923 & 1.838 & 2.429 & 1.863 & 2.013 \\ 
        EMOCA & 1.868 & 2.679 & 2.426 & 1.438 & 2.103 \\ 
        MICA & 1.585 & 3.478 & 2.374 & 1.099 & 2.134 \\ 
        RingNet & 1.934 & 2.074 & 2.995 & 2.028 & 2.258 \\ 
        
        \hline

        \textbf{Side View} & & & & & \\
        
        HRN & 1.642 & 1.285 & 1.906 & 1.038 & 1.468 \\ 
        Deep3D & 1.749 & 1.411 & 2.074 & 1.528 & 1.691 \\  
        \textbf{Our} & 1.576 & 2.218 & 2.142 & 1.112 & \textbf{1.762} \\  
        PRNet & 1.868 & 1.856 & 2.445 & 1.960 & 2.032 \\ 
        DECA & 1.903 & 2.472 & 2.423 & 1.630 & 2.107 \\ 
        EMOCA & 1.867 & 2.636 & 2.448 & 1.548 & 2.125 \\ 
        MICA & 1.525 & 2.636 & 2.448 & 1.548 & 2.125 \\ 
        RingNet & 1.921 & 1.994 & 3.081 & 2.027 & 2.256 \\ 
        
        \hline
        
    \end{tabular}}
    \vspace{-3mm}
    \label{comparision}
\end{table}


\vspace{-4mm}

\begin{table}[th]
\centering
\caption{Diversity values of randomly generated meshes for various methods. Higher is better. }
\vspace{-2mm}
\scalebox{0.85}{\begin{tabular}{c|c|c|c|c}
 -        &  Deep3D \cite{deep3dface_recon}   &   DECA \cite{deca}  &  MICA \cite{MICA:ECCV2022}  & Ours  \\
\hline
DIV &  0.18  &  1.13  & 0.21   &   \textbf{1.67}  
\label{table:div_spec} 
\end{tabular}}
\vspace{-2mm}
\end{table}


\subsubsection{Diversity metrics}

One of the important features of a good 3D face model is how diverse it's synthesized meshes are. Similar to prior works that generate 3D meshes~\cite{abrevaya2019decoupled, taherkhani2022controllable}, we measure global diversity as the mean vertex distance over all possible pairs of $n$ meshes. Table \ref{table:div_spec} reports the diversity values for $n=1000$ meshes synthesized by MICA\cite{MICA:ECCV2022}, DECA\cite{deca} and Deep3D\cite{deep3dface_recon}.

\begin{figure}[h]
\centering
\includegraphics[width=0.45\textwidth]{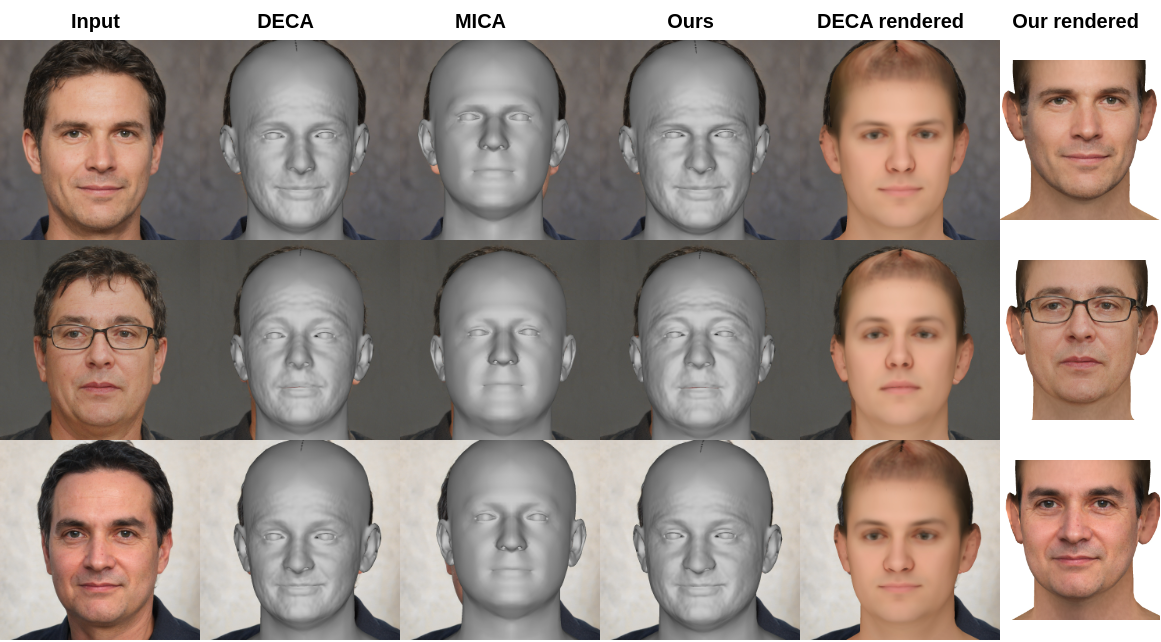}
\caption{3D Face Reconstruction comparison with DECA\cite{deca} \& MICA \cite{MICA:ECCV2022}. DECA is unable to model head shape accurately and provides a smooth texture. MICA only outputs a smooth, neutral and frontal mesh. Our method produces a more detailed mesh with photo-realistic texture, capturing wrinkles, expressions and head shape accurately.}
\label{fig:mesh_comparison}
\vspace{-8mm}
\end{figure}

As illustrated in Fig.~\ref{fig:mesh_comparison}, our model captures head shapes better than DECA model which produces similar-looking head shapes. We observe a significant improvement in diversity statistics over MICA that only predicts a smooth and neutral mesh and was trained on limited 3D data.

\subsubsection{Evaluation of Rendered Faces}

We evaluate the quality of texture by comparing qualitatively to a recently proposed method, OSTeC \cite{Gecer_2021_CVPR}. We also quantitatively compare texture and rendered faces by rendering it on a mesh with other methods, including LiftedGAN~\cite{shi2021lifting}, DECA~\cite{deca}, and OSTeC~\cite{Gecer_2021_CVPR}.

\begin{figure}[h]
\centering
\includegraphics[width=0.40\textwidth]{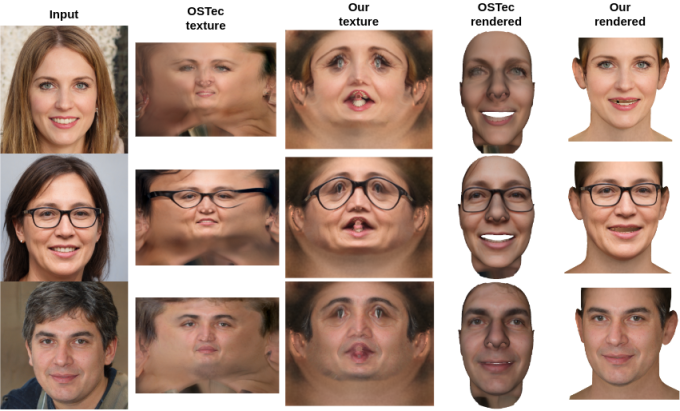}
\caption{Qualitative comparison of synthesized texture and rendered results with OSTeC \cite{Gecer_2021_CVPR}. 
OSTeC produces a smooth texture by stitching multiple images, often leading to artifacts.
Our approach can synthesize a better textured mesh outperforming OSTeC in preserving the details. }
\label{fig:OSTeC_vs_ours}
\vspace{-2mm}
\end{figure}

Fig. \ref{fig:OSTeC_vs_ours} shows the visual comparison between OSTeC \cite{Gecer_2021_CVPR} and our model.  OSTeC uses latent optimization based GAN inversion which adds random artifacts in the generated image like the beard appearing on some patches of the face. Furthermore, this leads to inconsistency while stitching textures across different poses.


For quantitative evaluation of our texture and rendered faces, we randomly sampled 1K images from a pretrained StyleGAN2 generator. We use the same set of $\textbf{z}$ to generate rendered 3D faces using our method. The corresponding 2D face images are used to perform 3D reconstruction using other methods\cite{deca, Gecer_2021_CVPR, shi2021lifting}. To compare how well the texture preserves identity, we measure identity similarity between the input image and the corresponding 3D faces rendered at multiple poses, including the original pose, $0$\textdegree, $\pm15$\textdegree, $\pm30$\textdegree, and $\pm45$\textdegree.  The observations can be found in Table \ref{table:id_similarity}. As inferred from the table, our method not only performs better in capturing the identity for the front poses but also for a large number of side poses. More evaluations can be found in supplementary section A.



\begin{table}[h]
\centering

\vspace{-1mm}
\scalebox{0.85}{\begin{tabular}{c|c|c|c|c|c}
 Method & same pose & $0$\textdegree & $\pm 15$\textdegree & $\pm 30$\textdegree & $\pm 45$\textdegree \\
 \hline
 LiftedGAN \cite{shi2021lifting} & 0.90 & 0.88 & 0.854 & 0.83 & 0.804 \\
 DECA \cite{deca} & 0.869 & 0.799 & 0.76 & 0.69 & 0.63 \\
 OSTeC \cite{Gecer_2021_CVPR} & 0.952 & 0.939 & 0.921 & 0.906 & 0.88 \\
 Ours & \textbf{0.999} & \textbf{0.995} & \textbf{0.988} & \textbf{0.972} & \textbf{0.941} 
\vspace{-1mm}
\label{table:id_similarity} 
\end{tabular}}
\caption{ID similarity comparison between the input image and the corresponding 3D face rendered at various poses.}
\vspace{-4mm}
\end{table}

\begin{figure}[h]
\centering
\includegraphics[width=0.40\textwidth]{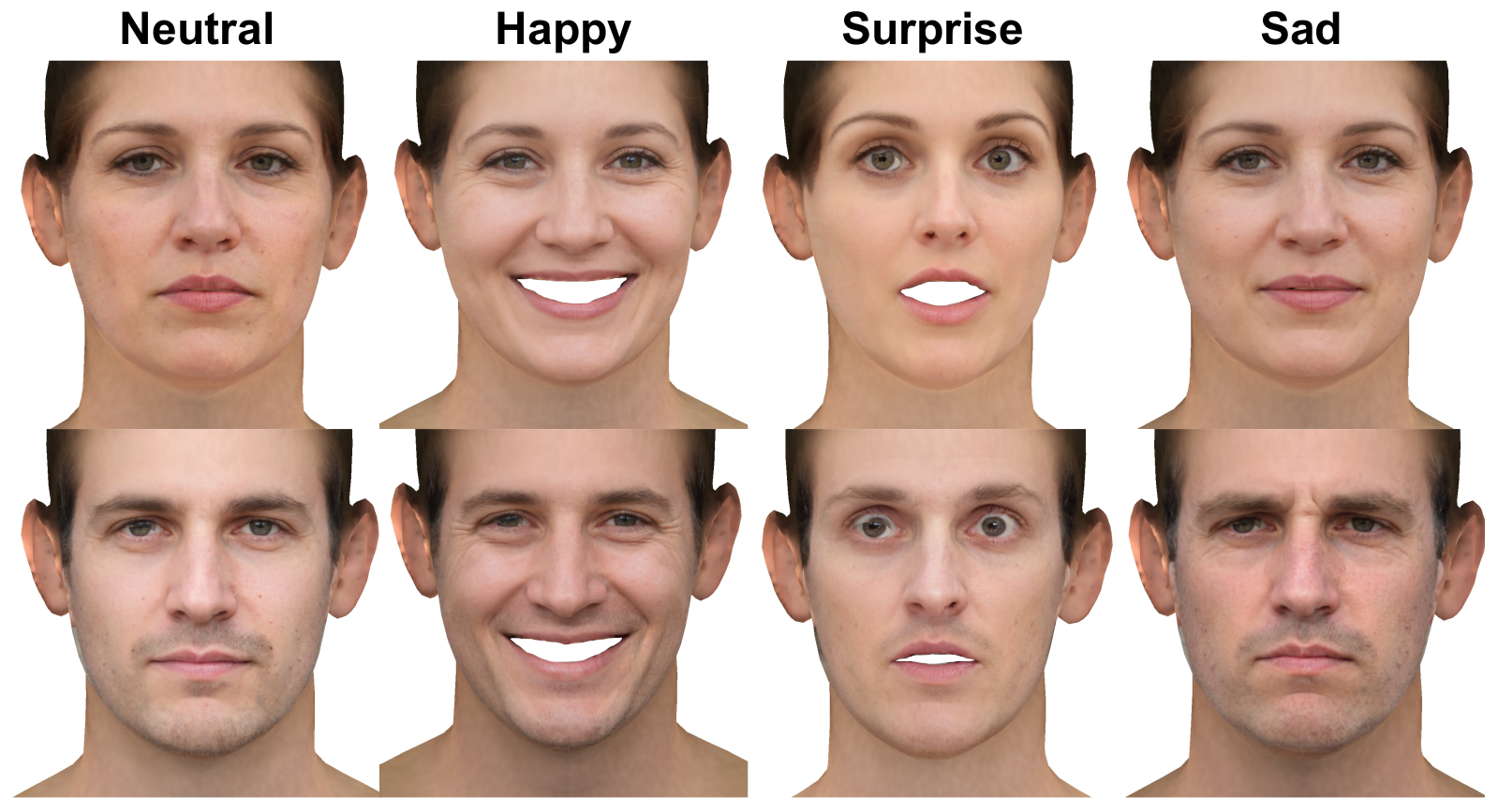}
\caption{Generating different expressions by manipulating $\textbf{w} \in \mathbb{R}^{18 \times 512}$. The first column in each row is the neutral expression corresponding to randomly sampled $\textbf{w}$. The subsequent columns show different expressions for the same identity by varying $\textbf{w}$. }
\label{fig:expressions}
\vspace{-4mm}
\end{figure}

\subsection{3D face manipulation}

 Once our end-to-end pipeline is trained and is able to generate albedo and mesh corresponding to a $\textbf{z} \in \mathbb{R}^{512}$ or $\textbf{w} \in \mathbb{R}^{18 \times 512}$, it opens up an avenue for latent space manipulations in 3D rendered faces. This section shows examples of editing 3D faces directly from the latent space or text.

{\bf Latent space manipulation:} Given a latent code $\textbf{w} \in \mathbb{R}^{18 \times 512}$, we can get the modified latent code $\textbf{w}_{edit} \in \mathbb{R}^{18 \times 512}$ by $\textbf{w}_{edit} = \textbf{w} + \alpha \textbf{n} $, where $\textbf{n} \in \mathbb{R}^{512}$ is a vector orthogonal to the semantic boundary and $\alpha \in \mathbb{R}$ is a constant. More details on latent space manipulations using semantic boundaries can be found in \cite{shen2020interfacegan, rai2021improved}. Fig. \ref{fig:expressions} shows the result of latent space manipulations for generating different expressions of the same identity. 

{\bf Text based 3D Face Editing:} Similar to latent space manipulations, we can perform text-based 3D face editing using Contrastive Language-Image Pretraining
(CLIP) models~\cite{radford2021learning}. We used StyleCLIP \cite{patashnik2021styleclip} model to enable text-based editing. The text-guided latent optimization and latent residual mapper strategies of StyleCLIP are implemented in $\textbf{w} \in \mathbb{R}^{18 \times 512}$ space and make it easy for us to plug into our pipeline. This paper shows results for the text-guided latent optimization approach, which can be trained for random text queries. Given a $\textbf{w}$ for a face, StyleCLIP can produce an updated latent code $\textbf{w}' \in \mathbb{R}^{18 \times 512}$ corresponding to a given text query and input $\textbf{w}$. Fig. \ref{fig:fig1}(c) illustrates some sample results generated for the text-based 3D face editing.  Refer to supplementary Section B.5 for more results on text-based editing.


\section{Conclusion and Future Work}


In this paper, we attempt to develop a high-quality 3D face generation pipeline. We propose AlbedoGAN that synthesizes albedo and generalizes well over multiple poses capturing intrinsic details of the face. Our approach generates meshes that capture high-frequency details like face wrinkles. Comprehensive experiments demonstrate superiority of our method over others in predicting detailed mesh and preserving the identity in reconstructed 3D faces. 
As a consequence of using StyleGAN2 based pipeline, we bring style editing, semantic face manipulations, and text-based editing in 3D faces.


While our pipeline can generate high-quality 3D faces from StyleGAN2's latent space, some issues still need to be addressed. Mesh-based representations are unable to model details like hair. We foresee exploiting topologically uniform mesh, and a NeRF-based approach should be able to capture such facial features. We'll extend our work to incorporate more complex illumination models.



{\small
\bibliographystyle{ieee_fullname}
\bibliography{iccv_paper_review}

\begin{thebibliography}{10}\itemsep=-1pt

\bibitem{abrevaya2019decoupled}
Victoria~Fern{\'a}ndez Abrevaya, Adnane Boukhayma, Stefanie Wuhrer, and Edmond
  Boyer.
\newblock A decoupled 3d facial shape model by adversarial training.
\newblock In {\em Proceedings of the IEEE/CVF International Conference on
  Computer Vision}, pages 9419--9428, 2019.

\bibitem{latent_space2}
Yuval Alaluf, Or Patashnik, and Daniel Cohen-Or.
\newblock Only a matter of style: Age transformation using a style-based
  regression model.
\newblock {\em ACM Transactions on Graphics (TOG)}, 40(4):1--12, 2021.

\bibitem{3dmm}
Volker Blanz and Thomas Vetter.
\newblock A morphable model for the synthesis of 3d faces, 1999.

\bibitem{facewarehouse}
Chen Cao, Yanlin Weng, Shun Zhou, Yiying Tong, and Kun Zhou.
\newblock Facewarehouse: A 3d facial expression database for visual computing.
\newblock {\em IEEE Transactions on Visualization and Computer Graphics},
  20(3):413--425, 2013.

\bibitem{pi_gan}
Eric~R. Chan, Marco Monteiro, Petr Kellnhofer, Jiajun Wu, and Gordon Wetzstein.
\newblock Pi-gan: Periodic implicit generative adversarial networks for
  3d-aware image synthesis.
\newblock In {\em Proceedings of the IEEE/CVF Conference on Computer Vision and
  Pattern Recognition (CVPR)}, pages 5799--5809, June 2021.

\bibitem{LYHM}
Hang Dai, Nick Pears, William Smith, and Christian Duncan.
\newblock Statistical modeling of craniofacial shape and texture.
\newblock {\em International Journal of Computer Vision}, 128(2):547--571,
  2020.

\bibitem{uvgan}
Jiankang Deng, Shiyang Cheng, Niannan Xue, Yuxiang Zhou, and Stefanos
  Zafeiriou.
\newblock Uv-gan: Adversarial facial uv map completion for pose-invariant face
  recognition.
\newblock In {\em Proceedings of the IEEE conference on computer vision and
  pattern recognition}, pages 7093--7102, 2018.

\bibitem{deng2018arcface}
Jiankang Deng, Jia Guo, Xue Niannan, and Stefanos Zafeiriou.
\newblock Arcface: Additive angular margin loss for deep face recognition.
\newblock In {\em CVPR}, 2019.

\bibitem{deng2019arcface}
Jiankang Deng, Jia Guo, Niannan Xue, and Stefanos Zafeiriou.
\newblock Arcface: Additive angular margin loss for deep face recognition.
\newblock In {\em Proceedings of the IEEE/CVF conference on computer vision and
  pattern recognition}, pages 4690--4699, 2019.

\bibitem{deep3dface_recon}
Yu Deng, Jiaolong Yang, Sicheng Xu, Dong Chen, Yunde Jia, and Xin Tong.
\newblock Accurate 3d face reconstruction with weakly-supervised learning: From
  single image to image set.
\newblock In {\em IEEE Computer Vision and Pattern Recognition Workshops},
  2019.

\bibitem{unrealengine}
{Epic Games}.
\newblock Unreal engine.

\bibitem{HRN}
Biwen et al.
\newblock A hierarchical representation network for accurate and detailed face
  reconstruction from in-the-wild images.
\newblock In {\em CVPR}, 2023.

\bibitem{deep3d_rebuttal}
Yu et al.
\newblock Accurate 3d face reconstruction with weakly-supervised learning: From
  single image to image set.
\newblock In {\em CVPRW}, 2019.

\bibitem{REALY}
Zenghao et al.
\newblock Realy: Rethinking the evaluation of 3d face reconstruction.
\newblock In {\em ECCV}, 2022.

\bibitem{deca}
Yao Feng, Haiwen Feng, Michael~J. Black, and Timo Bolkart.
\newblock Learning an animatable detailed {3D} face model from in-the-wild
  images.
\newblock In {\em ACM Transactions on Graphics, (Proc. SIGGRAPH)}, volume~40,
  2021.

\bibitem{stirling}
Zhen-Hua Feng, Patrik Huber, Josef Kittler, Peter Hancock, Xiao-Jun Wu, Qijun
  Zhao, Paul Koppen, and Matthias R{\"a}tsch.
\newblock Evaluation of dense 3d reconstruction from 2d face images in the
  wild.
\newblock In {\em 2018 13th IEEE International Conference on Automatic Face \&
  Gesture Recognition (FG 2018)}, pages 780--786. IEEE, 2018.

\bibitem{nerface}
Guy Gafni, Justus Thies, Michael Zollhofer, and Matthias Nie{\ss}ner.
\newblock Dynamic neural radiance fields for monocular 4d facial avatar
  reconstruction.
\newblock In {\em Proceedings of the IEEE/CVF Conference on Computer Vision and
  Pattern Recognition}, pages 8649--8658, 2021.

\bibitem{Gecer_2021_CVPR}
Baris Gecer, Jiankang Deng, and Stefanos Zafeiriou.
\newblock Ostec: One-shot texture completion.
\newblock In {\em Proceedings of the IEEE/CVF Conference on Computer Vision and
  Pattern Recognition (CVPR)}, pages 7628--7638, June 2021.

\bibitem{gecer2020synthesizing}
Baris Gecer, Alexandros Lattas, Stylianos Ploumpis, Jiankang Deng, Athanasios
  Papaioannou, Stylianos Moschoglou, and Stefanos Zafeiriou.
\newblock Synthesizing coupled 3d face modalities by trunk-branch generative
  adversarial networks.
\newblock In {\em European conference on computer vision}, pages 415--433.
  Springer, 2020.

\bibitem{gecer2020tbgan}
Baris {Gecer}, Alexander {Lattas}, Stylianos {Ploumpis}, Jiankang {Deng},
  Athanasios {Papaioannou}, Stylianos {Moschoglou}, and Stefanos {Zafeiriou}.
\newblock Synthesizing coupled 3d face modalities by trunk-branch generative
  adversarial networks.
\newblock In {\em Proceedings of the European conference on computer vision
  (ECCV)}, 2020.

\bibitem{Gecer_2019_CVPR}
Baris Gecer, Stylianos Ploumpis, Irene Kotsia, and Stefanos Zafeiriou.
\newblock Ganfit: Generative adversarial network fitting for high fidelity 3d
  face reconstruction.
\newblock In {\em The IEEE Conference on Computer Vision and Pattern
  Recognition (CVPR)}, June 2019.

\bibitem{gecer2019ganfit}
Baris Gecer, Stylianos Ploumpis, Irene Kotsia, and Stefanos Zafeiriou.
\newblock Ganfit: Generative adversarial network fitting for high fidelity 3d
  face reconstruction.
\newblock In {\em Proceedings of the IEEE/CVF conference on computer vision and
  pattern recognition}, pages 1155--1164, 2019.

\bibitem{fastganfit}
Baris Gecer, Stylianos Ploumpis, Irene Kotsia, and Stefanos Zafeiriou.
\newblock Fast-ganfit: Generative adversarial network for high fidelity 3d face
  reconstruction.
\newblock {\em arXiv preprint arXiv:2105.07474}, 2021.

\bibitem{neural-head-avatars}
Philip-William Grassal, Malte Prinzler, Titus Leistner, Carsten Rother,
  Matthias Nie{\ss}ner, and Justus Thies.
\newblock Neural head avatars from monocular rgb videos.
\newblock In {\em Proceedings of the IEEE/CVF Conference on Computer Vision and
  Pattern Recognition}, pages 18653--18664, 2022.

\bibitem{guo2021sample}
Jia Guo, Jiankang Deng, Alexandros Lattas, and Stefanos Zafeiriou.
\newblock Sample and computation redistribution for efficient face detection.
\newblock {\em arXiv preprint arXiv:2105.04714}, 2021.

\bibitem{he2016deep}
Kaiming He, Xiangyu Zhang, Shaoqing Ren, and Jian Sun.
\newblock Deep residual learning for image recognition.
\newblock In {\em Proceedings of the IEEE conference on computer vision and
  pattern recognition}, pages 770--778, 2016.

\bibitem{ichim2015dynamic}
Alexandru~Eugen Ichim, Sofien Bouaziz, and Mark Pauly.
\newblock Dynamic 3d avatar creation from hand-held video input.
\newblock {\em ACM Transactions on Graphics (ToG)}, 34(4):1--14, 2015.

\bibitem{karras2020analyzing}
Tero Karras, Samuli Laine, Miika Aittala, Janne Hellsten, Jaakko Lehtinen, and
  Timo Aila.
\newblock Analyzing and improving the image quality of stylegan.
\newblock In {\em Proceedings of the IEEE/CVF conference on computer vision and
  pattern recognition}, pages 8110--8119, 2020.

\bibitem{stylegan2}
Tero Karras, Samuli Laine, Miika Aittala, Janne Hellsten, Jaakko Lehtinen, and
  Timo Aila.
\newblock Analyzing and improving the image quality of stylegan.
\newblock In {\em Proceedings of the IEEE/CVF conference on computer vision and
  pattern recognition}, pages 8110--8119, 2020.

\bibitem{kim2018deep}
Hyeongwoo Kim, Pablo Garrido, Ayush Tewari, Weipeng Xu, Justus Thies, Matthias
  Niessner, Patrick P{\'e}rez, Christian Richardt, Michael Zollh{\"o}fer, and
  Christian Theobalt.
\newblock Deep video portraits.
\newblock {\em ACM Transactions on Graphics (TOG)}, 37(4):1--14, 2018.

\bibitem{texture-completion-iccv}
Jongyoo Kim, Jiaolong Yang, and Xin Tong.
\newblock Learning high-fidelity face texture completion without complete face
  texture.
\newblock In {\em Proceedings of the IEEE/CVF International Conference on
  Computer Vision}, pages 13990--13999, 2021.

\bibitem{occlusion_robust_mofa}
Chunlu Li, Andreas Morel-Forster, Thomas Vetter, Bernhard Egger, and Adam
  Kortylewski.
\newblock To fit or not to fit: Model-based face reconstruction and occlusion
  segmentation from weak supervision.
\newblock In {\em arXiv preprint arXiv:2106.09614}, 2021.

\bibitem{flame}
Tianye Li, Timo Bolkart, Michael.~J. Black, Hao Li, and Javier Romero.
\newblock Learning a model of facial shape and expression from {4D} scans.
\newblock {\em ACM Transactions on Graphics, (Proc. SIGGRAPH Asia)}, 36(6),
  2017.

\bibitem{lombardi2018deep}
Stephen Lombardi, Jason Saragih, Tomas Simon, and Yaser Sheikh.
\newblock Deep appearance models for face rendering.
\newblock {\em ACM Transactions on Graphics (ToG)}, 37(4):1--13, 2018.

\bibitem{Nichol2021GLIDETP}
Alex Nichol, Prafulla Dhariwal, Aditya Ramesh, Pranav Shyam, Pamela Mishkin,
  Bob McGrew, Ilya Sutskever, and Mark Chen.
\newblock {GLIDE}: Towards photorealistic image generation and editing with
  text-guided diffusion models.
\newblock {\em ICML}, 2022.

\bibitem{olivier2021facetunegan}
Nicolas Olivier, Kelian Baert, Fabien Danieau, Franck Multon, and Quentin
  Avril.
\newblock Facetunegan: Face autoencoder for convolutional expression transfer
  using neural generative adversarial networks.
\newblock {\em arXiv preprint arXiv:2112.00532}, 2021.

\bibitem{paszke2019pytorch}
Adam Paszke, Sam Gross, Francisco Massa, Adam Lerer, James Bradbury, Gregory
  Chanan, Trevor Killeen, Zeming Lin, Natalia Gimelshein, Luca Antiga, et~al.
\newblock Pytorch: An imperative style, high-performance deep learning library.
\newblock {\em Advances in neural information processing systems}, 32, 2019.

\bibitem{patashnik2021styleclip}
Or Patashnik, Zongze Wu, Eli Shechtman, Daniel Cohen-Or, and Dani Lischinski.
\newblock Styleclip: Text-driven manipulation of stylegan imagery.
\newblock In {\em Proceedings of the IEEE/CVF International Conference on
  Computer Vision}, pages 2085--2094, 2021.

\bibitem{bfm}
Pascal Paysan, Reinhard Knothe, Brian Amberg, Sami Romdhani, and Thomas Vetter.
\newblock A 3d face model for pose and illumination invariant face recognition.
\newblock In {\em 2009 Sixth IEEE International Conference on Advanced Video
  and Signal Based Surveillance}, pages 296--301, 2009.

\bibitem{radford2021learning}
Alec Radford, Jong~Wook Kim, Chris Hallacy, Aditya Ramesh, Gabriel Goh,
  Sandhini Agarwal, Girish Sastry, Amanda Askell, Pamela Mishkin, Jack Clark,
  et~al.
\newblock Learning transferable visual models from natural language
  supervision.
\newblock In {\em International Conference on Machine Learning}, pages
  8748--8763. PMLR, 2021.

\bibitem{radford2019language}
Alec Radford, Jeffrey Wu, Rewon Child, David Luan, Dario Amodei, Ilya
  Sutskever, et~al.
\newblock Language models are unsupervised multitask learners.
\newblock {\em OpenAI blog}, 1(8):9, 2019.

\bibitem{rai2021improved}
Aashish Rai, Clara Ducher, and Jeremy~R Cooperstock.
\newblock Improved attribute manipulation in the latent space of stylegan for
  semantic face editing.
\newblock In {\em 2021 20th IEEE International Conference on Machine Learning
  and Applications (ICMLA)}, pages 38--43. IEEE, 2021.

\bibitem{ramesh2022hierarchical}
Aditya Ramesh, Prafulla Dhariwal, Alex Nichol, Casey Chu, and Mark Chen.
\newblock Hierarchical text-conditional image generation with clip latents.
\newblock {\em arXiv preprint arXiv:2204.06125}, 2022.

\bibitem{lolnerf}
Daniel Rebain, Mark Matthews, Kwang~Moo Yi, Dmitry Lagun, and Andrea
  Tagliasacchi.
\newblock Lolnerf: Learn from one look.
\newblock In {\em Proceedings of the IEEE/CVF Conference on Computer Vision and
  Pattern Recognition}, pages 1558--1567, 2022.

\bibitem{Saharia2022PhotorealisticTD}
Chitwan Saharia, William Chan, Saurabh Saxena, Lala Li, Jay Whang, Emily~L.
  Denton, Seyed Kamyar~Seyed Ghasemipour, Burcu~Karagol Ayan, Seyedeh~Sara
  Mahdavi, Raphael~Gontijo Lopes, Tim Salimans, Jonathan Ho, David Fleet, and
  Mohammad Norouzi.
\newblock Photorealistic text-to-image diffusion models with deep language
  understanding.
\newblock {\em NeurIPS}, 2022.

\bibitem{ringnet}
Soubhik Sanyal, Timo Bolkart, Haiwen Feng, and Michael Black.
\newblock Learning to regress 3d face shape and expression from an image
  without 3d supervision.
\newblock In {\em Proceedings IEEE Conf. on Computer Vision and Pattern
  Recognition (CVPR)}, June 2019.

\bibitem{schroff2015facenet}
Florian Schroff, Dmitry Kalenichenko, and James Philbin.
\newblock Facenet: A unified embedding for face recognition and clustering.
\newblock In {\em Proceedings of the IEEE conference on computer vision and
  pattern recognition}, pages 815--823, 2015.

\bibitem{shen2020interfacegan}
Yujun Shen, Ceyuan Yang, Xiaoou Tang, and Bolei Zhou.
\newblock Interfacegan: Interpreting the disentangled face representation
  learned by gans.
\newblock In {\em TPAMI}, 2020.

\bibitem{shi2021lifting}
Yichun Shi, Divyansh Aggarwal, and Anil~K Jain.
\newblock Lifting 2d stylegan for 3d-aware face generation.
\newblock In {\em Proceedings of the IEEE/CVF Conference on Computer Vision and
  Pattern Recognition}, pages 6258--6266, 2021.

\bibitem{taherkhani2022controllable}
Fariborz Taherkhani, Aashish Rai, Quankai Gao, Shaunak Srivastava, Xuanbai
  Chen, Fernando de~la Torre, Steven Song, Aayush Prakash, and Daeil Kim.
\newblock Controllable 3d generative adversarial face model via disentangling
  shape and appearance.
\newblock {\em arXiv preprint arXiv:2208.14263}, 2022.

\bibitem{tewari}
Ayush Tewari, Michael Zollh{\"{o}}fer, Hyeongwoo Kim, Pablo Garrido, Florian
  Bernard, Patrick P{\'{e}}rez, and Christian Theobalt.
\newblock Mofa: Model-based deep convolutional face autoencoder for
  unsupervised monocular reconstruction.
\newblock {\em CoRR}, abs/1703.10580, 2017.

\bibitem{thies2016face2face}
Justus Thies, Michael Zollhofer, Marc Stamminger, Christian Theobalt, and
  Matthias Nie{\ss}ner.
\newblock Face2face: Real-time face capture and reenactment of rgb videos.
\newblock In {\em Proceedings of the IEEE conference on computer vision and
  pattern recognition}, pages 2387--2395, 2016.

\bibitem{toshpulatov2021generative}
Mukhiddin Toshpulatov, Wookey Lee, and Suan Lee.
\newblock Generative adversarial networks and their application to 3d face
  generation: a survey.
\newblock {\em Image and Vision Computing}, 108:104119, 2021.

\bibitem{unity}
{Unity Technologies}.
\newblock Unity.

\bibitem{weise2011realtime}
Thibaut Weise, Sofien Bouaziz, Hao Li, and Mark Pauly.
\newblock Realtime performance-based facial animation.
\newblock {\em ACM transactions on graphics (TOG)}, 30(4):1--10, 2011.

\bibitem{textureGeneration}
Xiangnan Yin, Di Huang, Zehua Fu, Yunhong Wang, and Liming Chen.
\newblock Weakly-supervised photo-realistic texture generation for 3d face
  reconstruction.
\newblock {\em arXiv preprint arXiv:2106.08148}, 2021.

\bibitem{yin2020fan}
Xi Yin, Ying Tai, Yuge Huang, and Xiaoming Liu.
\newblock Fan: Feature adaptation network for surveillance face recognition and
  normalization.
\newblock In {\em Proceedings of the Asian Conference on Computer Vision},
  2020.

\bibitem{latent_space1}
Yin Yu, Ghasedi Kamran, Wu HsiangTao, Yang Jiaolong, Tong Xi, and Fu Yun.
\newblock Expanding the latent space of stylegan for real face editing.
\newblock {\em arXiv preprint arXiv:2204.12530}, 2022.

\bibitem{imavatar}
Yufeng Zheng, Victoria~Fern{\'a}ndez Abrevaya, Marcel~C B{\"u}hler, Xu Chen,
  Michael~J Black, and Otmar Hilliges.
\newblock Im avatar: Implicit morphable head avatars from videos.
\newblock In {\em Proceedings of the IEEE/CVF Conference on Computer Vision and
  Pattern Recognition}, pages 13545--13555, 2022.

\bibitem{mofa_nerf}
Yiyu Zhuang, Hao Zhu, Xusen Sun, and Xun Cao.
\newblock Mofanerf: Morphable facial neural radiance field.
\newblock {\em arXiv preprint arXiv:2112.02308}, 2021.

\bibitem{MICA:ECCV2022}
Wojciech Zielonka, Timo Bolkart, and Justus Thies.
\newblock Towards metrical reconstruction of human faces.
\newblock In {\em European Conference on Computer Vision}, 2022.

\end{thebibliography}
}

\end{document}